%% file: iclr2025_conference.tex
\documentclass{article} 
\usepackage{iclr2025_conference,times,graphicx}
\usepackage[export]{adjustbox}

\input{math_commands.tex}

\usepackage{hyperref}
\usepackage{url}
\usepackage{graphicx}
\usepackage{subcaption}
\usepackage{arydshln}
\usepackage{placeins}

\title{Zero-Shot Tree Detection and Segmentation from Aerial Forest Imagery}

\author{Michelle Chen \textsuperscript{1} \hspace{1mm} David Russell \textsuperscript{2} \hspace{1mm} Amritha Pallavoor \textsuperscript{2} \hspace{1mm} Derek Young \textsuperscript{2} \hspace{1mm} Jane Wu \textsuperscript{1}\\
\textsuperscript{1} UC Berkeley \hspace{1mm}\textsuperscript{2} UC Davis\\
\texttt{\textsuperscript{1} \{michelle.chenn,janehwu\}@berkeley.edu}\\
\texttt{\textsuperscript{2} \{djrussell,aspallavoor,djyoung\}@ucdavis.edu}
}

\iclrfinalcopy 
\begin{document}

\maketitle

\begin{abstract}
Large-scale delineation of individual trees from remote sensing imagery is crucial to the advancement of ecological research, particularly as climate change and other environmental factors rapidly transform forest landscapes across the world. Current RGB tree segmentation methods rely on training specialized machine learning models with labeled tree datasets. While these learning-based approaches can outperform manual data collection when accurate, the existing models still depend on training data that's hard to scale. In this paper, we investigate the efficacy of using a state-of-the-art image segmentation model, Segment Anything Model 2 (SAM2), in a \textit{zero-shot} manner for individual tree detection and segmentation. We evaluate a pretrained SAM2 model on two tasks in this domain: (1) zero-shot segmentation and (2) zero-shot transfer by using predictions from an existing tree detection model as prompts. Our results suggest that SAM2 not only has impressive generalization capabilities, but also can form a natural synergy with specialized methods trained on in-domain labeled data. We find that applying large pretrained models to problems in remote sensing is a promising avenue for future progress. We make our
code available at: \url{https://github.com/open-forest-observatory/tree-detection-framework}.
\end{abstract}

\section{Introduction}
Remote sensing of global forest landscapes plays a critical role in a broad range of research areas, from ecological monitoring~\citep{lechner2020applications,fassnacht2024remote} to wildfire prevention~\citep{iban2022machine} to climate change resiliency~\citep{falk2022mechanisms,forzieri2022emerging}.
In particular, the ability to identify and measure individual trees provides valuable insights into forest structure, biodiversity, carbon sequestration, and overall ecosystem health.
Traditional methods of forest inventory, such as ground-based surveys, are time-consuming, labor-intensive, and often limited in scale.
In contrast, remotely sensed overhead imagery (e.g.\ via unmanned aerial vehicles) combined with machine learning techniques offers a promising solution for large-scale tree analysis.


Existing machine learning methods for tree detection are CNN-based models that were trained or fine-tuned on manually labeled tree crown datasets.
DeepForest~\citep{weinstein2020deepforest} uses the RetinaNet architecture~\citep{lin2017focal} and retrains from the lidar-based method in \citet{silva2016imputation} using RGB imagery from sites in the National Ecological Observatory Network (NEON) dataset~\citep{weinstein2019individual}.
In a similar vein, Detectree2~\citep{ball2023accurate} introduced a Mask R-CNN model~\citep{he2017mask} based on Detectron2 and trained on images collected in Malaysia and French Guiana.
Because these approaches rely on models that were only trained on relatively small, labeled tree datasets, they can naturally struggle with generalization to unseen geographical areas.
Following the rise of transformer-based models~\citep{vaswani2017attention} trained on Internet-scale data, there is growing evidence that large pretrained models can be effectively applied in a zero-shot manner to various scientific domains (see e.g.\ \citet{cambrin2024depth,gutierrez2024no}).



Building on recent advancements in computer vision, we investigate the efficacy of using a pretrained Segment Anything Model 2 (SAM2)~\citep{ravi2024sam} in a zero-shot manner for automatic tree detection and segmentation from aerial imagery. Utilizing transformer-based architecture with proven zero-shot generalization capabilities provides significant benefits over previous methods, and our proposed approach could enhance generalization across diverse ecosystems around the world.
In order to evaluate the generalization capabilities of SAM2, we investigate (1) zero-shot prediction using a pretrained SAM2 model and (2) whether zero-shot transfer from tree detection to segmentation is viable by prompting SAM2 using bounding boxes predicted by DeepForest.
We benchmark pretrained SAM2 against DeepForest and Detectree2 using two standard datasets from NEON and Detectree2.
Furthermore, qualitative analysis of Emerald Point imagery from the Open Forest Observatory~\footnote{https://openforestobservatory.org}, which previously served as the basis for geometric tree detection optimization~\citep{young2022optimizing}, demonstrates that SAM2 significantly outperforms existing models.
Our experiments highlight the potential for large pretrained models in machine learning to drive progress in remote sensing, particularly for visual understanding at scale.


\section{Zero-Shot Prediction and Transfer}
\textbf{Segment Anything Model 2.} SAM2~\citep{ravi2024sam} is a foundation model designed for promptable visual segmentation across images and videos.
Given an input RGB image and a set of point, bounding box, or mask prompts, the model outputs segmentation masks conditioned on the prompts.
The neural network architectures for both SAM2 and its predecessor, SAM~\citep{kirillov2023segment}, consist of an encoder-decoder structure using transformers.
The image encoder is a vision transformer~\citep{ryali2023hiera} pretrained using Masked Autoencoders (MAE)~\citep{he2022masked}, and the mask decoder includes a stack of transformer blocks to update the image and prompt embeddings in a bidirectional manner.
In addition to performance enhancements on images, SAM2 extends SAM to the video domain to enable both segmentation and tracking across time.
At present, SAM2 represents the current state-of-the-art for general image and video segmentation.



\begin{figure*}[ht]
    \centering
    \begin{subfigure}[b]{0.24\linewidth}
    \centering
    \includegraphics[width=\linewidth]{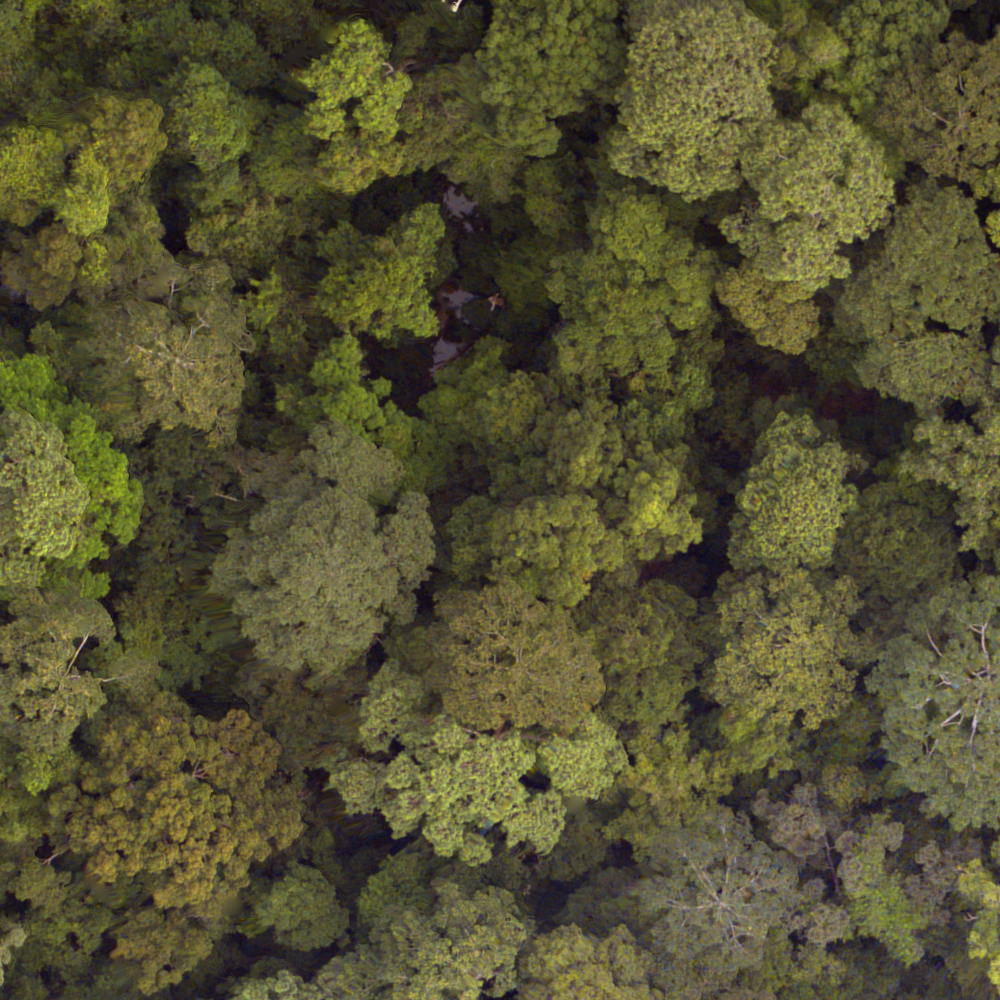}
    \caption{Input Image}
    \end{subfigure}
    \begin{subfigure}[b]{0.24\linewidth}
    \centering\
    \includegraphics[width=\linewidth,trim={5mm 5mm 5mm 5mm},clip]{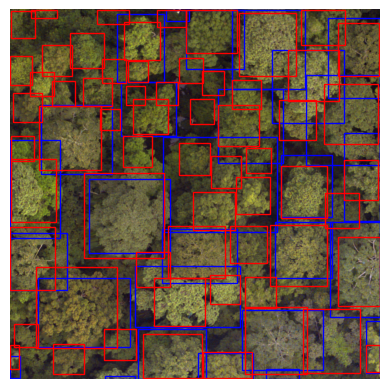}
    \caption{DeepForest}
    \end{subfigure}
    \begin{subfigure}[b]{0.24\linewidth}
    \centering
    \includegraphics[width=\linewidth,trim={5mm 5mm 5mm 5mm},clip]{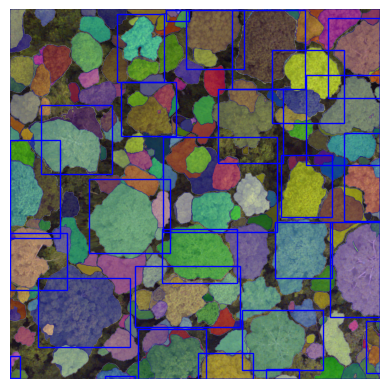}
    \caption{SAM2}
    \label{fig:zero_shot_sam2}
    \end{subfigure}
    \begin{subfigure}[b]{0.24\linewidth}
    \centering
    \includegraphics[width=\linewidth,trim={5mm 5mm 5mm 5mm},clip]{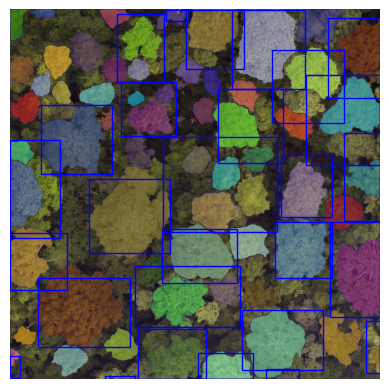}
    \caption{DeepForest + SAM2}
    \label{fig:zero_shot_df_sam2}
    \end{subfigure}
    \caption{SAM's prompt-based segmentation framework enables (c) \textit{zero-shot tree segmentation} and (d) \textit{zero-shot transfer} of tree detection models such as (b) DeepForest to SAM2 via bounding box prompting. An example image from the Detectree2 dataset is used. Ground truth bounding boxes are drawn in blue, predicted bounding boxes are in red, and predicted masks are randomly colored.}
    \vspace{-5mm}
\end{figure*}

\textbf{Zero-shot Tree Segmentation.}
We first aim to understand SAM2's zero-shot capabilities as a standalone method for automatic tree segmentation and detection.
Throughout our study, we use a pretrained SAM2 model with the Hiera-L image encoder and refer to this zero-shot prediction approach as ``SAM2.''
To segment trees across an image, we utilize SAM2's automatic mask generator, which uniformly samples single-point prompts in a grid structure that each yields a corresponding mask.
Due to the large-scale format of aerial imagery, we find that running the model on cropped sections and applying the standard post-processing to remove disconnected regions significantly improves segmentation quality.
After mask generation, bounding box predictions can be calculated for each mask to obtain tree detection results, which we use for quantitative evaluation on labeled tree crown data.
Example SAM2 zero-shot predictions are shown in Figure~\ref{fig:zero_shot_sam2}.

\textbf{Zero-shot Transfer from Tree Detection.}
SAM's prompt-based segmentation approach is designed for zero-shot transfer to downstream segmentation tasks. 
This means that through prompt engineering, knowledge from \textit{domain-specific models} can be directly transferred to SAM/SAM2.
In the context of vision-based tree delineation, the vast majority of existing models~\citep{fromm2019automated,weinstein2020deepforest,ball2023accurate,gan2023tree} are tree crown detection models that output either bounding boxes or polygons, which are much less labor-intensive to label than segmentation masks.
Aiming to build on this domain-specific expertise, we also investigate the efficacy of providing bounding box predictions from specialized tree detectors as prompts to SAM2.
Example SAM2 predictions using DeepForest bounding boxes are shown in Figure~\ref{fig:zero_shot_df_sam2}.

\section{Experiments}
\textbf{Datasets.} We evaluate SAM2's zero-shot capabilities on images drawn from three datasets spanning two continents: the Emerald Point dataset~\citep{young2022optimizing}, the National Ecological Observatory Network (NEON) TreeEvaluation dataset~\citep{weinstein2019individual,weinstein2021remote}, and the released Detectree2 dataset~\citep{ball2023accurate}.
First, in order to evaluate the generalization of both SAM2 and existing models, we use a dataset of aerial photographs captured by a quadcopter (referred to as Emerald Point) collected at a study site in Emerald Bay State Park located in California~\citep{young2022optimizing}.
For quantitative benchmarking, we use the NEON TreeEvaluation dataset, which includes 30,975 manually-annotated tree bounding boxes in 22 sites across the United States.
We also use the Danum, Sepilok East, and Sepilok West datasets released as part of Detectree2, which includes approximately 1,000 labeled crowns from two tropical field sites in Malaysia.



\textbf{Baselines.} We compare SAM2 with two CNN-based models for image-based tree detection: DeepForest~\citep{weinstein2020deepforest} and Detectree2~\citep{ball2023accurate}.
DeepForest was trained on NEON data, and Detectree2 was trained on the Danum, Sepilok, and Paracou sites in their collected dataset.
The specific models we use are the DeepForest 1.5.0 release and the Detectree2 2.0.1 release, both with default parameters.
For all methods, including SAM2, we apply non-maxmimum (NMS) suppression to predicted polygons with an intersection-over-union (IOU) threshold of 0.05. Applying NMS directly to polygon representations retains more detections compared to using bounding boxes, resulting in higher precision but lower recall values.
To the best of our knowledge, the data we evaluate all methods on was not used during training.

\begin{figure*}[ht]
    \centering
    \begin{subfigure}[b]{0.24\linewidth}
    \centering
    \includegraphics[width=\linewidth]{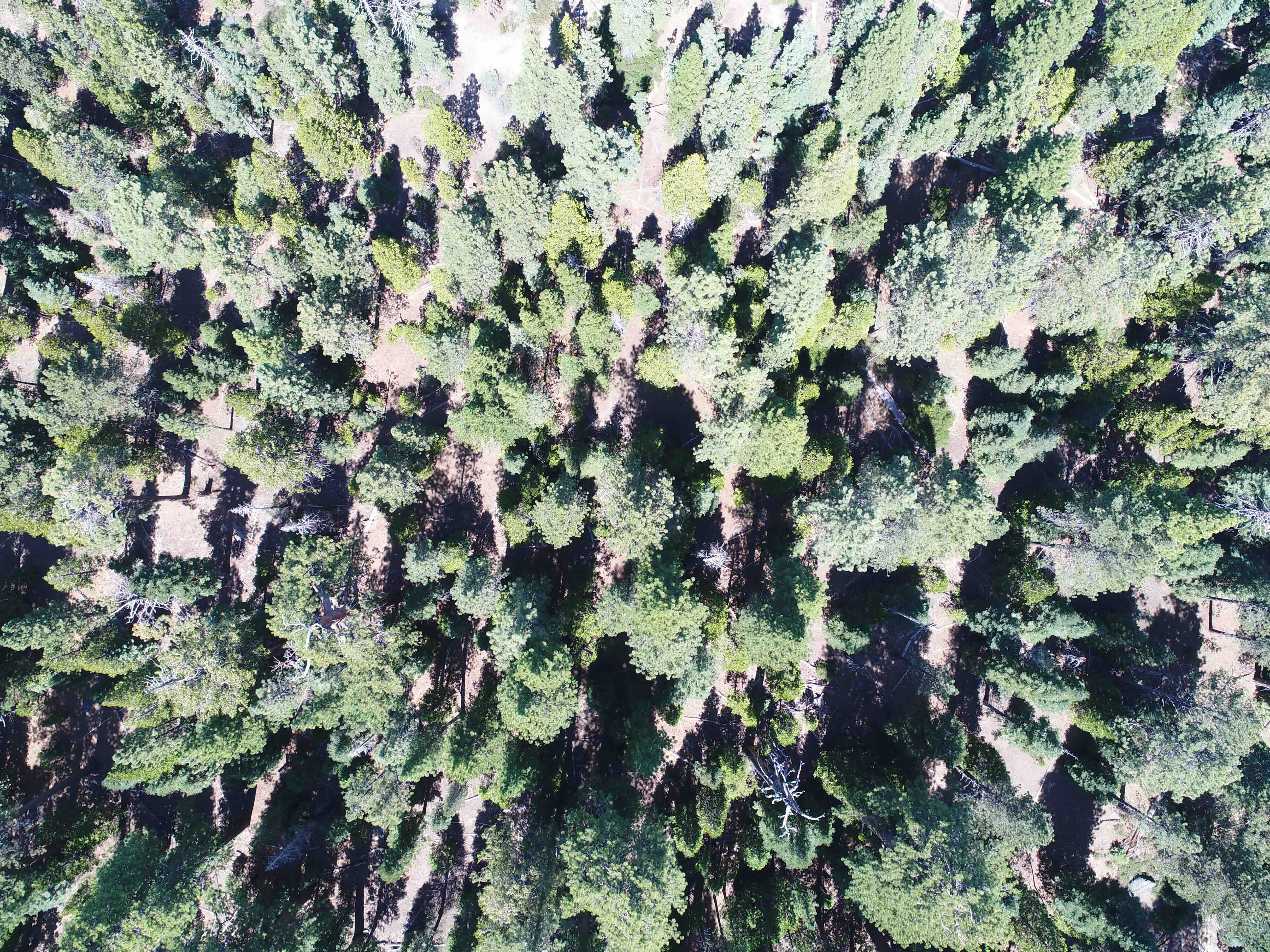}\\
    \includegraphics[width=\linewidth]{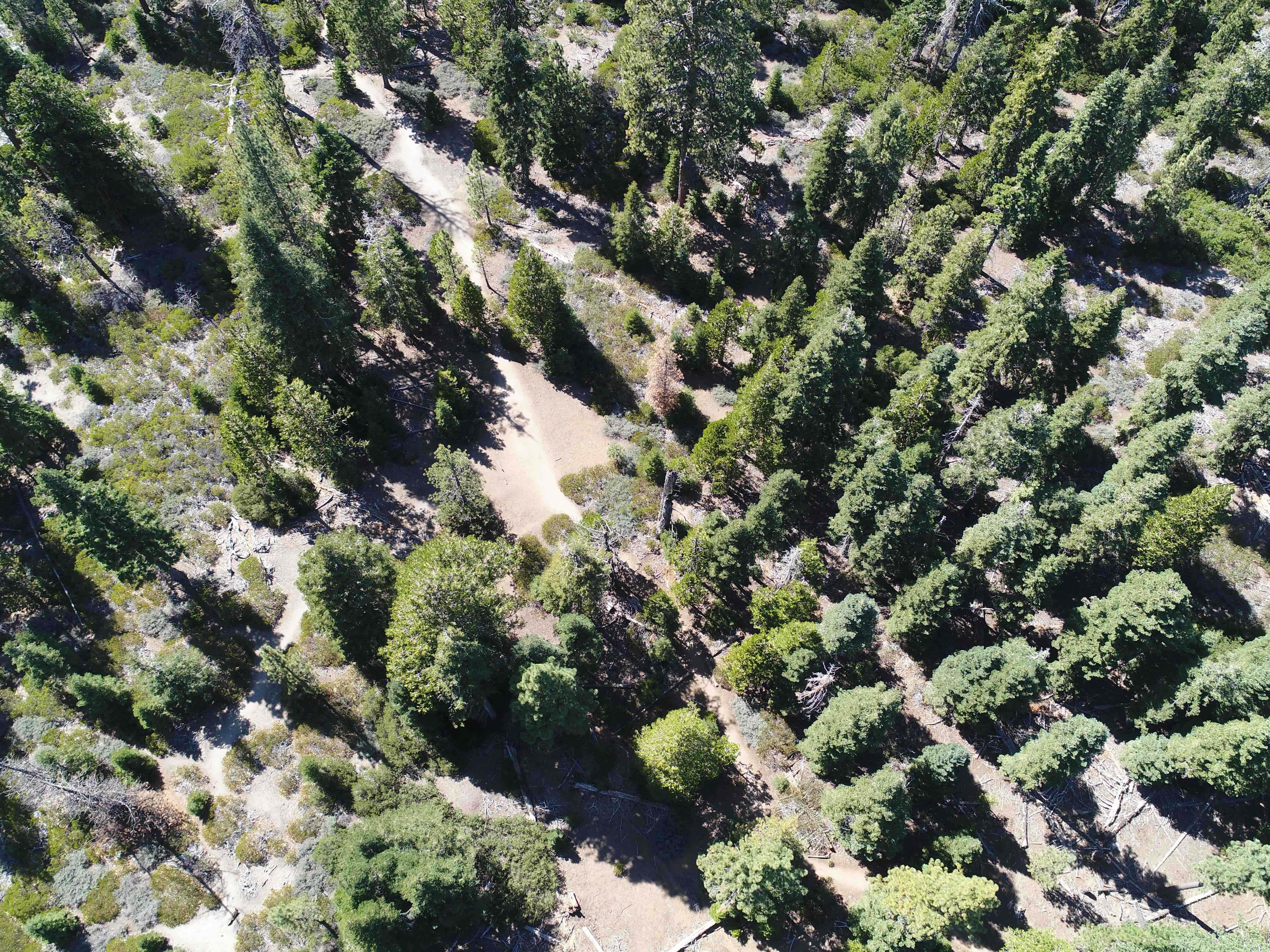}\\
    \includegraphics[width=\linewidth]{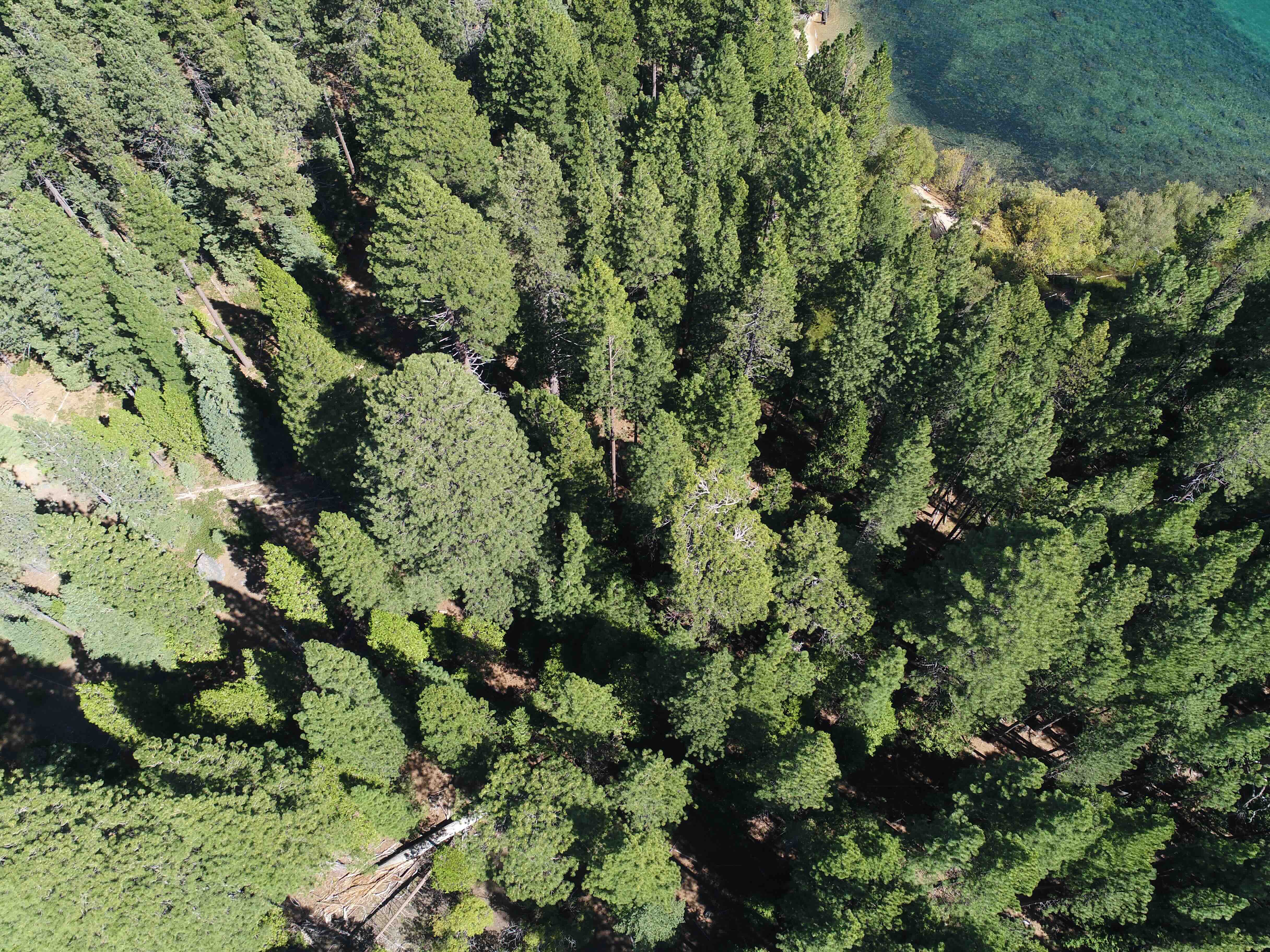}
    \caption{Input Image}
    \end{subfigure}
    \begin{subfigure}[b]{0.24\linewidth}
    \centering
    \includegraphics[width=\linewidth]{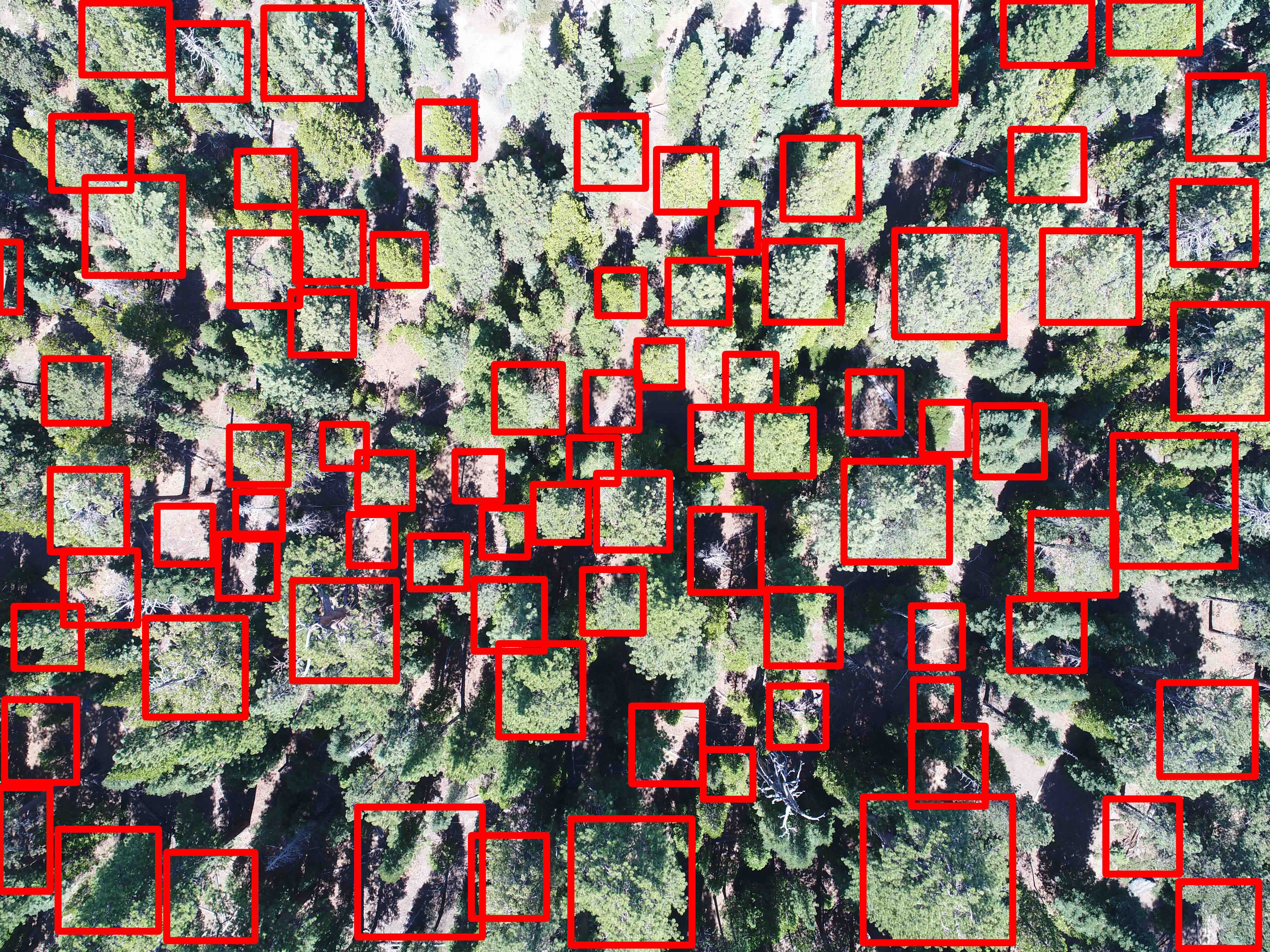}\\
    \includegraphics[width=\linewidth]{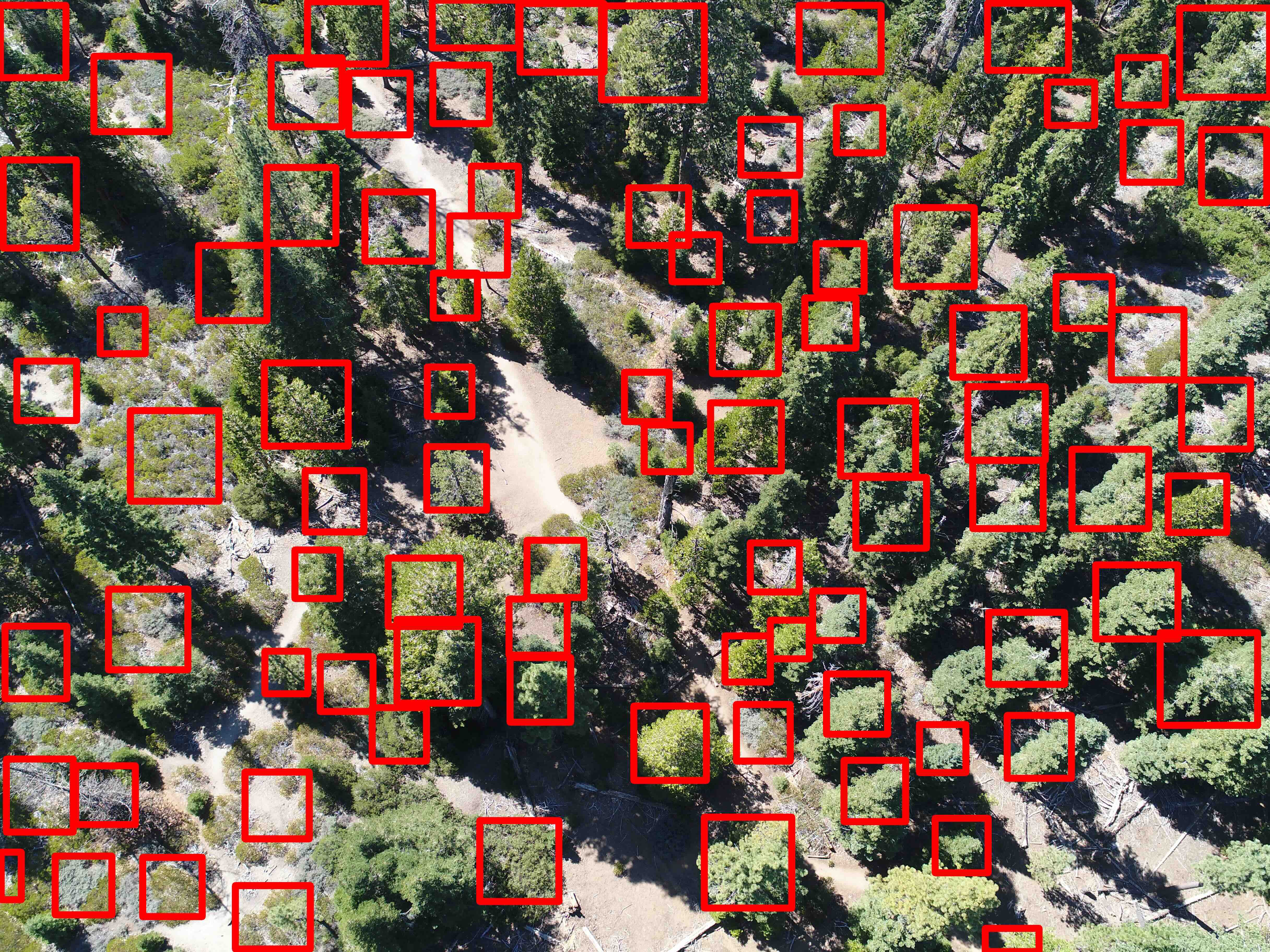}\\
    \includegraphics[width=\linewidth]{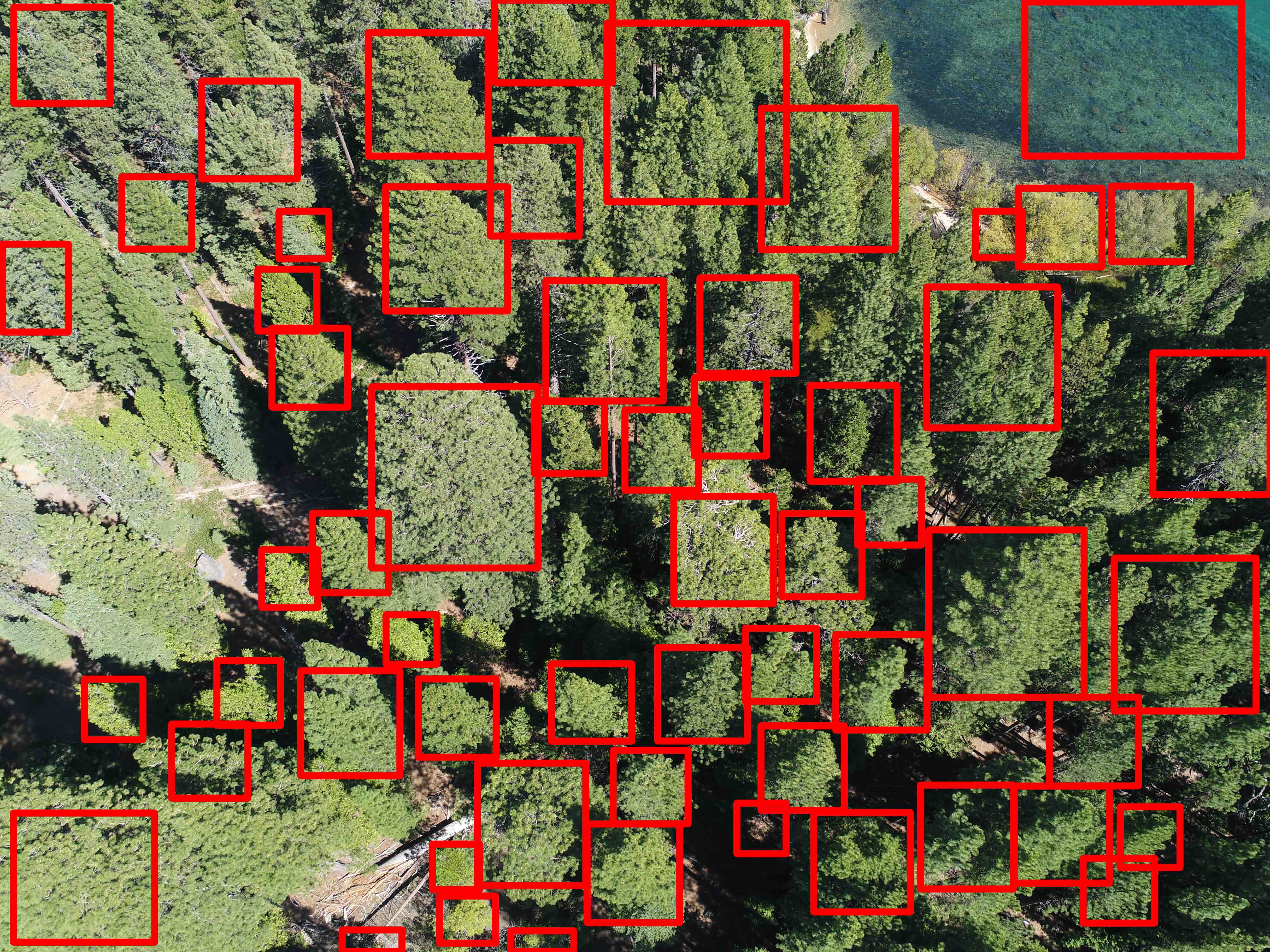}
    \caption{DeepForest}
    \end{subfigure}
    \begin{subfigure}[b]{0.24\linewidth}
    \centering
    \includegraphics[width=\linewidth]{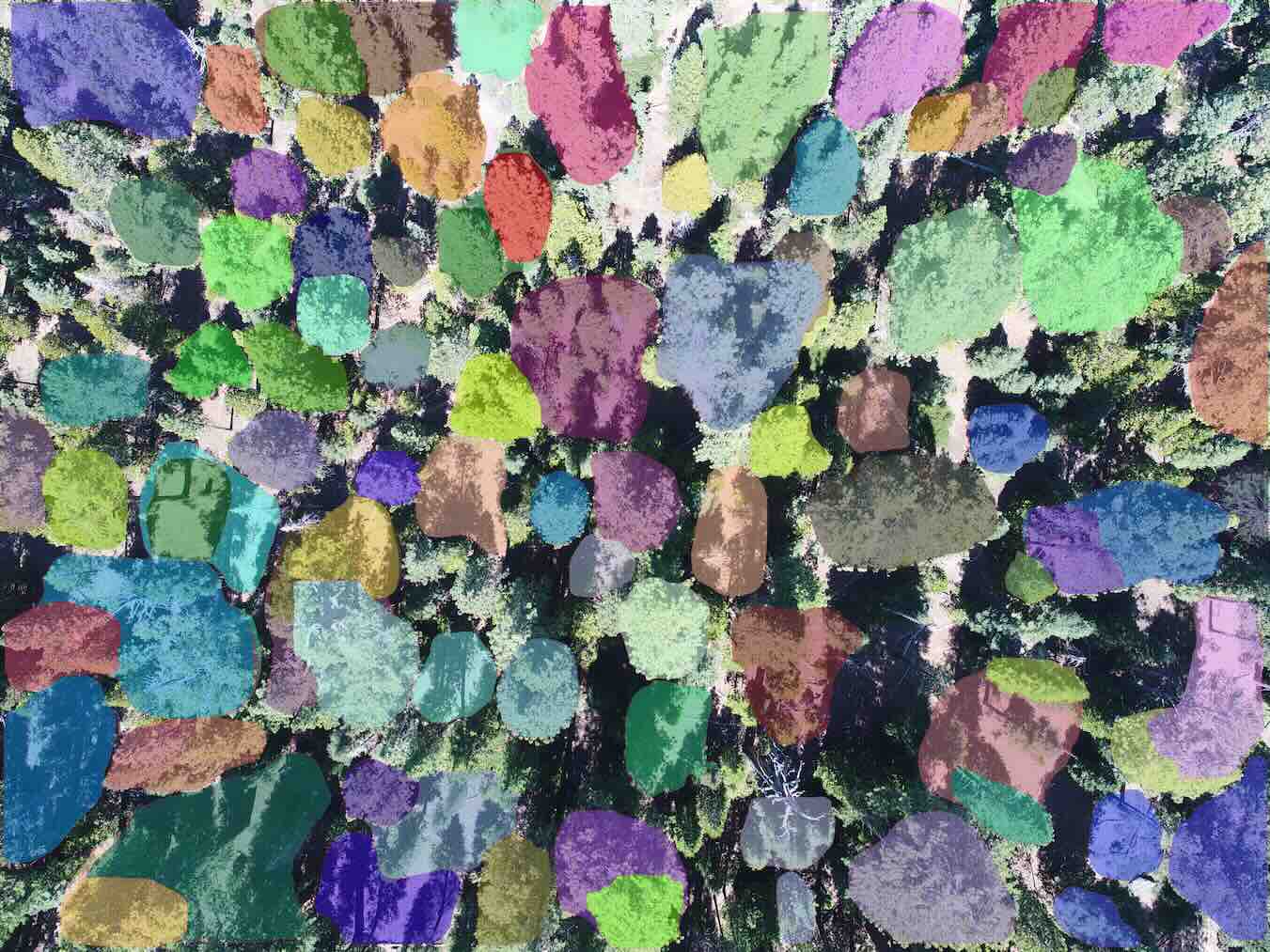}\\
    \includegraphics[width=\linewidth]{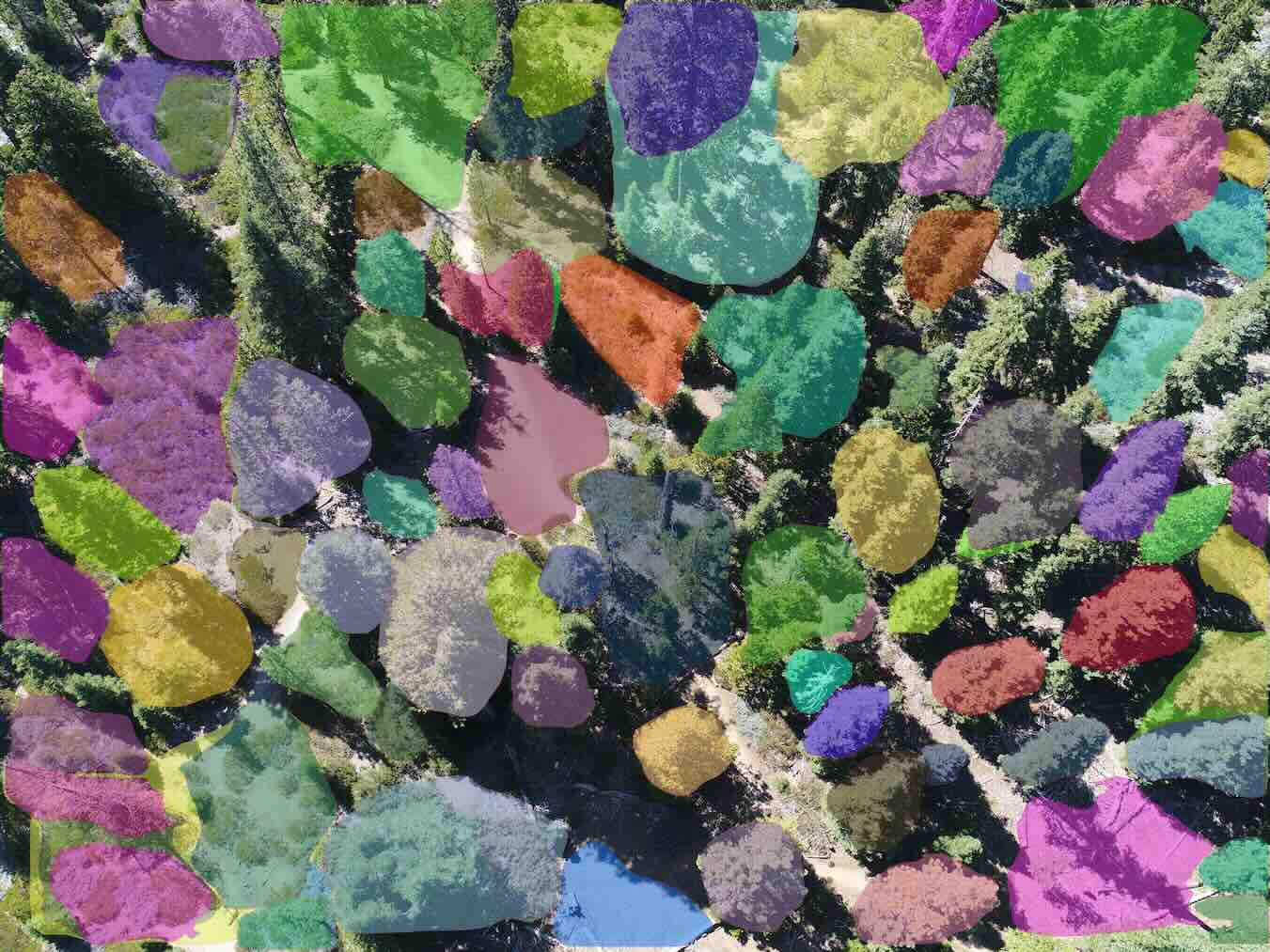}\\
    \includegraphics[width=\linewidth]{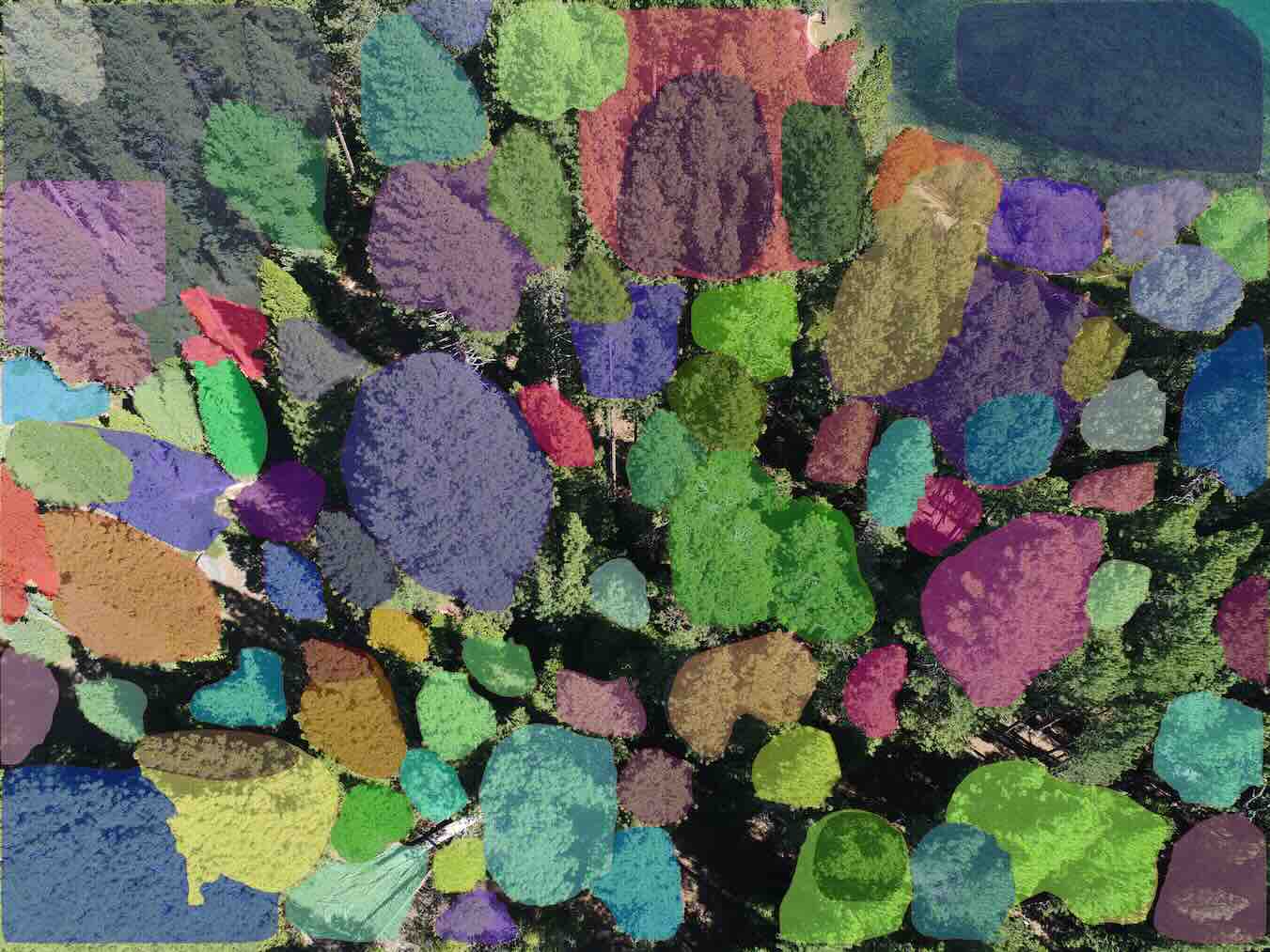}
    \caption{Detectree2}
    \end{subfigure}
    \begin{subfigure}[b]{0.24\linewidth}
    \centering
    \includegraphics[width=\linewidth]{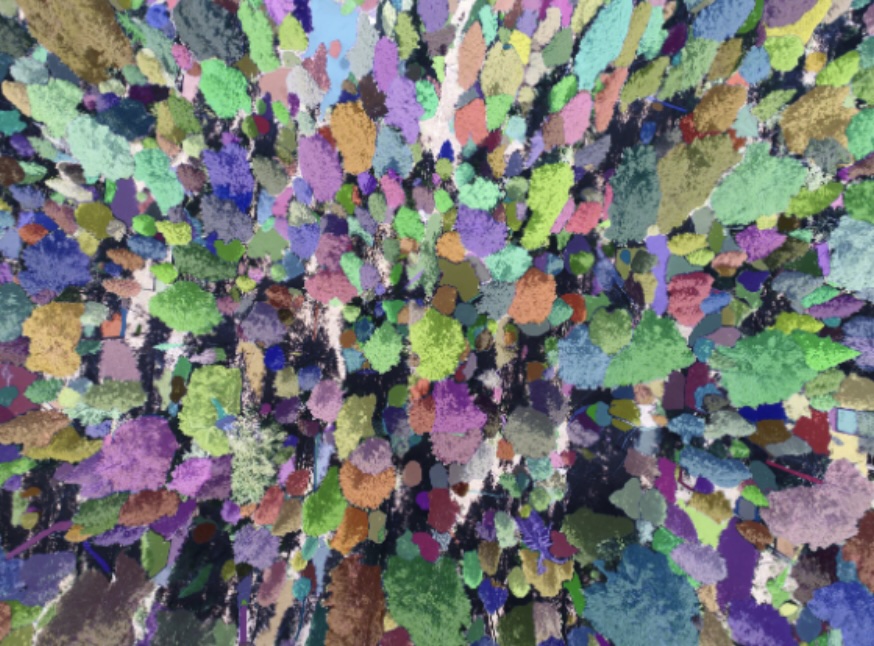}\\
    \includegraphics[width=\linewidth]{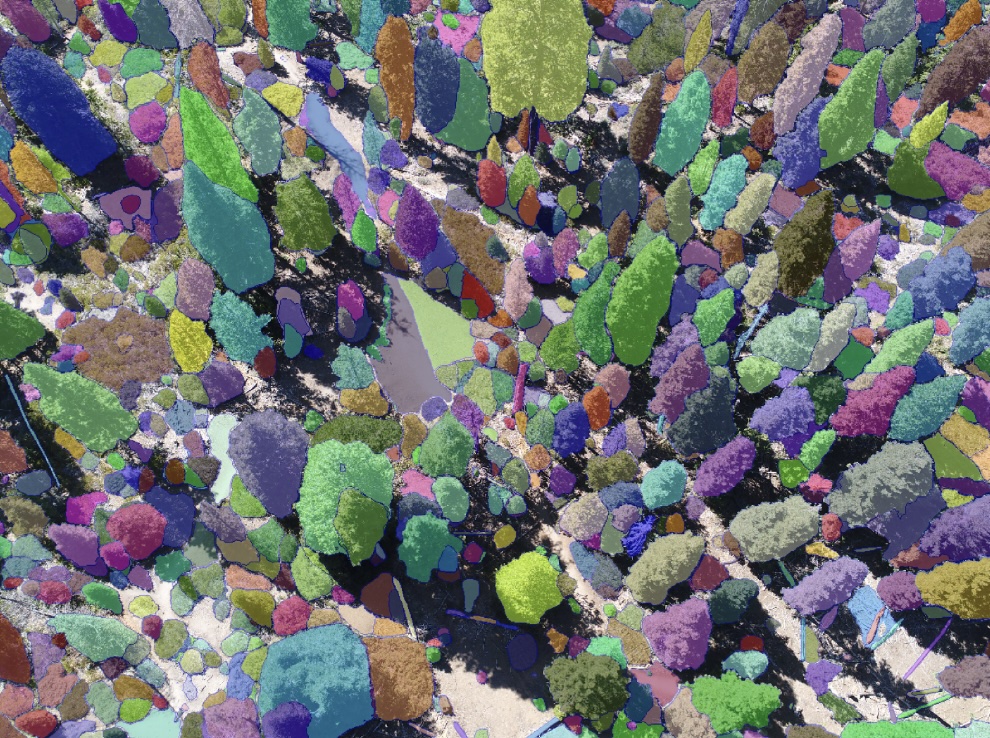}\\
    \includegraphics[width=\linewidth]{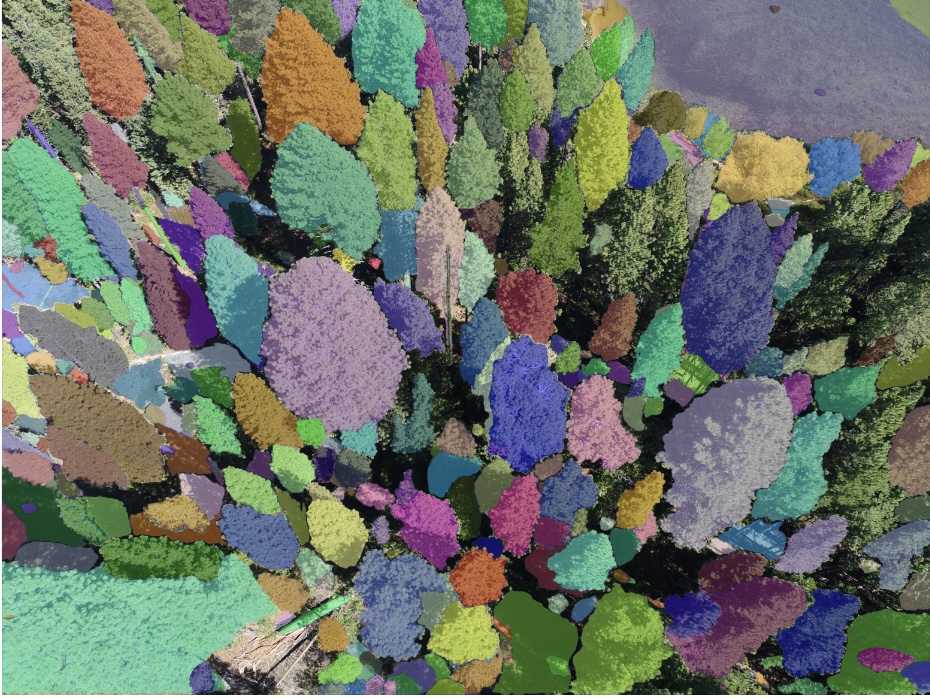}
    \caption{SAM2}
    \end{subfigure}
    \caption{Segmentation results using aerial images from the Emerald Point dataset. Each row includes the input image, DeepForest tree crown bounding boxes, Detectree2  polygons, and SAM2 segmentation masks.}
    \label{fig:ofo_results}
    \vspace{-4mm}
\end{figure*}

\textbf{Generalization to Emerald Point.} Neither DeepForest/Detectree2 nor SAM2 was trained using data from Emerald Bay State Park, so we can evaluate how well all three models generalize to unseen images/geographical regions using the Emerald Point dataset.
Figure~\ref{fig:ofo_results} shows qualitative comparisons between predicted DeepForest bounding boxes, Detectree2 polygons, and SAM2 polygon masks generated using automatic mask generation.
SAM2 is able to capture smaller and shorter trees with high accuracy, offering more detailed and reliable segmentation across varied forest densities.
Though DeepForest outputs bounding boxes and Detectree2 outputs polygons, the number and quality of detected trees are significantly lower than what SAM2 predicts. 
The drop in performance of existing models could be due in part to a change in camera perspective, as existing approaches were trained only on top-down orthomosaics. In contrast, SAM2 demonstrates the ability to effectively perform zero-shot segmentation from oblique perspectives in addition to top-down views.
While more extensive evaluations are needed, our results are an early indication that a pretrained SAM2 already shows impressive generalization capabilities to tree segmentation.



\begin{figure*}[ht]
    \centering
    \begin{subfigure}[b]{0.19\linewidth}
    \centering
    \includegraphics[width=\linewidth]{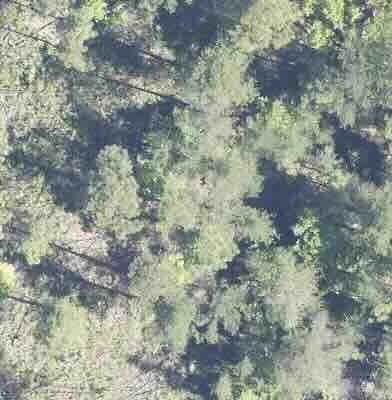}\\
    \includegraphics[width=\linewidth]{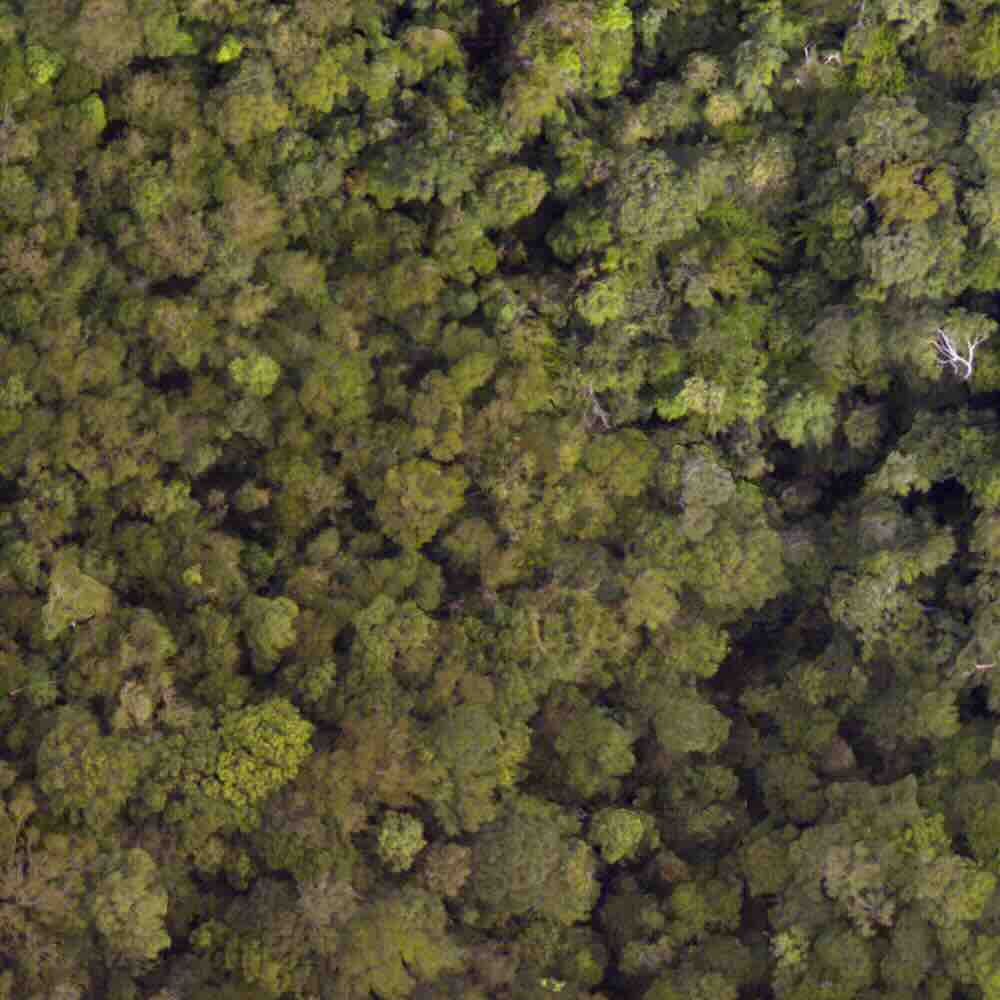}
    \caption{Input Image}
    \end{subfigure}
    \begin{subfigure}[b]{0.19\linewidth}
    \centering
    \includegraphics[width=\linewidth,trim={5mm 5mm 5mm 5mm},clip]{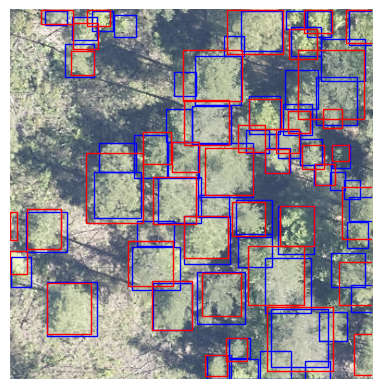}\\
    \includegraphics[width=\linewidth,trim={5mm 5mm 5mm 5mm},clip]{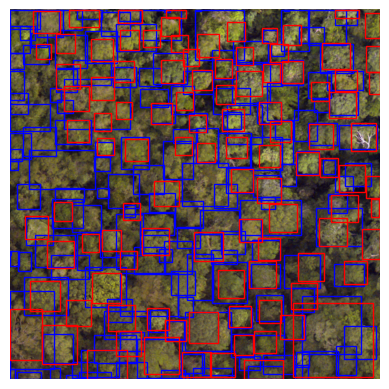}
    \caption{DeepForest}
    \end{subfigure}
    \begin{subfigure}[b]{0.19\linewidth}
    \centering
    \includegraphics[width=\linewidth,trim={5mm 5mm 5mm 5mm},clip]{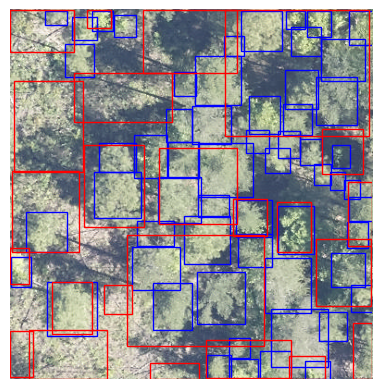}\\
    \includegraphics[width=\linewidth,trim={5mm 5mm 5mm 5mm},clip]{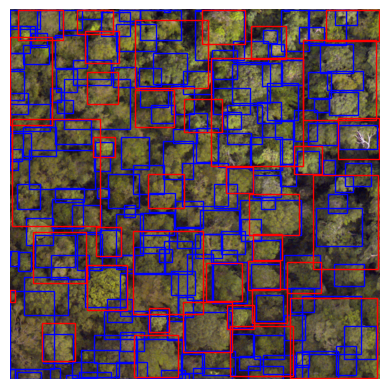}
    \caption{Detectree2}
    \end{subfigure}
    \begin{subfigure}[b]{0.19\linewidth}
    \centering
    \includegraphics[width=\linewidth,trim={5mm 5mm 5mm 5mm},clip]{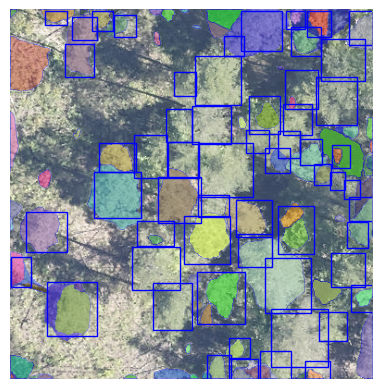}\\
    \includegraphics[width=\linewidth,trim={5mm 5mm 5mm 5mm},clip]{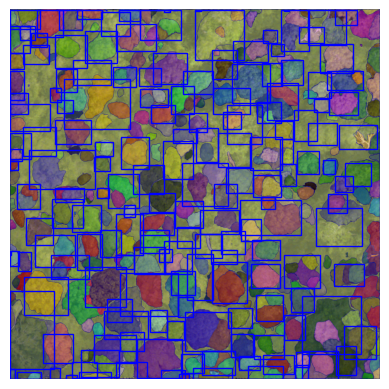}
    \caption{SAM2}
    \end{subfigure}
    \begin{subfigure}[b]{0.19\linewidth}
    \centering
    \includegraphics[width=\linewidth,trim={5mm 5mm 5mm 5mm},clip]{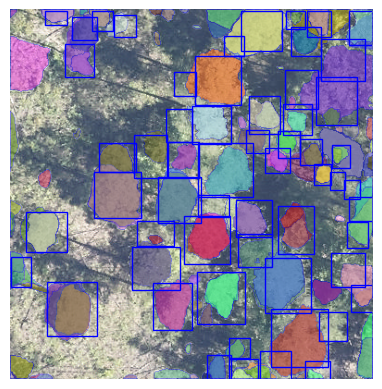}\\
    \includegraphics[width=\linewidth,trim={5mm 5mm 5mm 5mm},clip]{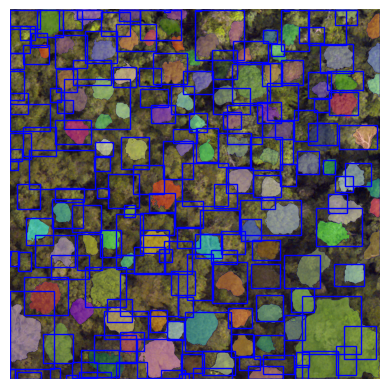}
    \caption{\scriptsize{DeepForest + SAM2}}
    \end{subfigure}
    \caption{Tree crown detection results using RGB images from the NEON (Row 1) and Detectree2 (Row 2) datasets. Ground truth bounding boxes are drawn in blue, and predictions are drawn in red. Each row includes the input image, DeepForest, Detectree2, and SAM2, and the bounding box prompted SAM2. Note that SAM2 tends to over-segment in the absence of prompts, which contributes to the low precision values in Table~\ref{tab:deeptree_comp}.}
    \label{fig:benchmark_comp}
    \vspace{-5mm}
\end{figure*}

\textbf{Benchmarking on tree crown datasets.}
While SAM2 is not technically an object detection model, we nonetheless provide some quantitative evaluation of its tree delineation capabilities by benchmarking on the NEON TreeEvaluation and Detectree2 datasets.
Table~\ref{tab:deeptree_comp} compares SAM2's zero-shot tree detection with DeepForest and Detectree2, both of which are performing in-distribution inference for their corresponding datasets (NEON and Detectree2).
For each dataset, we compute average precision and recall at an IOU threshold of 0.4 to evaluate the quality of tree crown bounding box predictions, with a minimum confidence for detections threshold of 0.1 (note that this does not exactly match how prior work metrics are reported).
Because SAM2 is (1) not meaningfully prompted and (2) does not distinguish between object classes, we observe that the model tends to over segment, which results in the low precision scores.
However, the results also indicate that SAM2 can achieve comparable recall to existing methods, particularly Detectree2, across both datasets.
Figure~\ref{fig:benchmark_comp} shows representative results from both the NEON and Detectree2 datasets.



\begin{table}[ht]
    \centering
    \begin{tabular}{c|cc|cc}
    \hline
    \multicolumn{1}{c}{} & \multicolumn{2}{c}{NEON}  & \multicolumn{2}{c}{Detectree2} \\
     Method  & Precision ($\uparrow$) & Recall ($\uparrow$) & Precision ($\uparrow$)  & Recall ($\uparrow$) \\
     \hline
     DeepForest & \textbf{0.61} & \textbf{0.76} & 0.34 & 0.60 \\
     Detectree2 & 0.25 & 0.50 & \textbf{0.40} & \textbf{0.72} \\
     SAM2 & 0.17 & \underline{0.41} & 0.16 & \underline{0.65} \\
    \end{tabular}
    \caption{Evaluation of tree crown bounding box detection on the NEON and Detectree2 benchmark datasets using DeepForest, Detectree2, and a pretrained SAM2 model (zero-shot). For all methods, bounding box precision and recall are determined using an IoU threshold of 0.4 and computed independently for each image before averaging across all images.}
    \label{tab:deeptree_comp}
    \vspace{-4mm}
\end{table}

\FloatBarrier

\section{Discussion}
In this paper, we explore the zero-shot capabilities of SAM2 for tree delineation in aerial remote sensing imagery.
The existing paradigm for tree detection and segmentation is to train specialized models using labeled tree crown data; however, we conjecture that recent progress in foundation models like SAM/SAM2 can greatly enhance the generalization capabilities of vision-based techniques in remote sensing.
We focus on two specific tasks in the domain of tree delineation: (1) zero-shot tree segmentation using SAM2 and (2) zero-shot transfer of tree detection to SAM2 segmentation. 
Our experimental results show strong indications that SAM2 has the potential to generalize to a wide range of tree species, canopy structures, and environmental conditions, perhaps even without any dataset-specific adaptation or fine-tuning.


\section*{Acknowledgements}
We would like to thank Marissa Ramirez de Chanlatte, Trevor Darrell, and Arjun Rewari for helpful discussions and feedback.
This work was supported in part by the National Science Foundation Division of Biological Infrastructure (\#2152671, \#2152672, \#2152673).
J. W. was supported by the NSF
Mathematical Sciences Postdoctoral Fellowship and the UC
President's Postdoctoral Fellowship.

\bibliography{iclr2025_conference}
\bibliographystyle{iclr2025_conference}


\end{document}

%% file: math_commands.tex
\usepackage{amsmath,amsfonts,bm}

\def\eqref#1{equation~\ref{#1}}

\def\1{\bm{1}}

\DeclareMathAlphabet{\mathsfit}{\encodingdefault}{\sfdefault}{m}{sl}
\SetMathAlphabet{\mathsfit}{bold}{\encodingdefault}{\sfdefault}{bx}{n}

